\documentclass[letterpaper, 10 pt, conference]{ieeeconf}
\IEEEoverridecommandlockouts
\usepackage{graphics}
\usepackage{amsmath}
\usepackage{amssymb} 
\usepackage{mathtools}
\usepackage{breqn}
\usepackage[framemethod=tikz]{mdframed}
\usepackage{subfigure}
\usepackage{booktabs}
\usepackage{hyperref}
\usepackage{xfrac}

\usepackage[math]{cellspace}
\usepackage{caption}
\definecolor{redish}{rgb}{0.74, 0.36, 0.36}
\definecolor{light-redish}{rgb}{0.88, 0.67, 0.67}

\title{\LARGE \bf
 Single-Stage Optimization of Open-loop Stable Limit Cycles with \\ Smooth, Symbolic Derivatives
}

\author{Muhammad Saud Ul Hassan$^1$ and Christian Hubicki$^1$ \thanks{$^1$Muhammad Saud Ul Hassan and Christian Hubicki are with the Department of Mechanical Engineering, FAMU-FSU College of Engineering, \texttt{\href{mailto:ms18ig@my.fsu.edu}{ms18ig@my.fsu.edu}},\, \texttt{\href{mailto:hubicki@eng.famu.fsu.edu}{hubicki@eng.famu.fsu.edu}}}}

\begin{document}

\begin{minipage}{\textwidth}\ \\[12pt]
  \centering
  © 2025 IEEE. Personal use of this material is permitted. \\[1em]
  Permission from IEEE must be obtained for all other uses, in any current or future media, including reprinting/republishing this material for advertising or promotional purposes, creating new collective works, for resale or redistribution to servers or lists, or reuse of any copyrighted component of this work in other works. \\[1em]
  DOI: \href{10.1109/ICRA55743.2025.11128720}{https://ieeexplore.ieee.org/document/11128720}
\end{minipage}

\newpage

\maketitle
\thispagestyle{empty}
\pagestyle{empty}

\begin{abstract}
Open-loop stable limit cycles are foundational to legged robotics, providing inherent self-stabilization that minimizes the need for computationally intensive feedback-based gait correction. While previous methods have primarily targeted specific robotic models, this paper introduces a general framework for rapidly generating limit cycles across various dynamical systems, with the flexibility to impose arbitrarily tight stability bounds. We formulate the problem as a single-stage constrained optimization problem and use Direct Collocation to transcribe it into a nonlinear program with closed-form expressions for constraints, objectives, and their gradients. 

Our method supports multiple stability formulations. In particular, we tested two popular formulations for limit cycle stability in robotics: (1) based on the spectral radius of a discrete return map, and (2) based on the spectral radius of the monodromy matrix, and tested five different constraint-satisfaction formulations of the eigenvalue problem to bound the spectral radius. We compare the performance and solution quality of the various formulations on a robotic swing-leg model, highlighting the Schur decomposition of the monodromy matrix as a method with broader applicability due to weaker assumptions and stronger numerical convergence properties. 

As a case study, we apply our method on a hopping robot model, generating open-loop stable gaits in under 2 seconds on an \href{https://ark.intel.com/content/www/us/en/ark/products/88195/intel-core-i76700k-processor-8m-cache-up-to-4-20-ghz.html}{Intel\textsuperscript\textregistered Core i7-6700K}, while simultaneously minimizing energy consumption even under tight stability constraints. 

\end{abstract}

%

\section{Introduction}
\textit{Open-loop stability} is the property of a system to naturally recover from perturbations. An open-loop stable walking robot, for instance, can traverse rough terrain with reduced dependence on feedback mechanisms for gait correction \cite{coleman1998stability}\cite{mombaur2000stable}\cite{mombaur2001human}, thereby reducing computational load, energy consumption, and the potential effects of latency in feedback systems that can make a robot’s motion less smooth and responsive. Despite this, most research in gait control has focused exclusively on feedback methods for stabilizing limit cycles \cite{westervelt2018feedback}, often neglecting control synthesis fusing open-loop stability with feedback methods \cite{wensing2023optimization}. We hypothesize that this omission is largely due to the challenges in reliably and efficiently computing open-loop stable limit cycles. In response, we present a general constrained optimization framework for rapidly generating open-loop stable limit cycles, with support for explicitly specified, arbitrarily tight stability constraints.




Ringrose \cite{ringrose1997self} was the first to achieve open-loop stable hopping using a robot with a large circular foot. Following this, Ernst et al. \cite{Ernst2009SpringLeggedLO} developed an open-loop deadbeat controller for a running monoped with a point foot. In contrast, other research has focused on optimization-based approaches that generalize to a wider class of systems. Grizzle et. al. \cite{898695} demonstrated an optimization method for systems with a single degree of underactuation. Mombaur et al. \cite{mombaur2001human}\cite{mombaur2005open}\cite{mombaur2005self}\cite{mombaur2006performing} developed a two-stage optimization strategy for finding open-loop stable controllers for specific robot models, where the inner loop searched for a periodic gait given the robot’s design parameters, and the outer loop determined the design parameters that stabilized the system. Later, Mombaur et al. \cite{mombaur2009using} proposed a single-loop optimization strategy using \textit{direct shooting}, though this method, which numerically computes integrals, is computationally intensive for underactuated systems. Dai et al. \cite{dai2012optimizing} proposed a trajectory optimization approach that positions the post-ground-impact states of a robot near a stable limit cycle, subsequently using feedback control to achieve convergence onto it.

\begin{figure}
    \centering
    \includegraphics[scale=1.0]{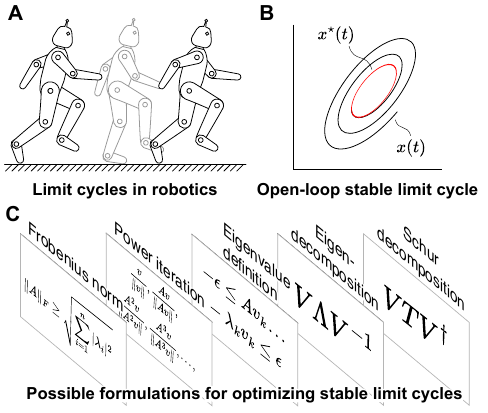}
    \caption{\textbf{A.} Cyclical motions are commonplace in robotics, particularly legged robotics. \textbf{B.} The stability of such cyclical motions ($x(t)$) can be casted in terms of limit cycle stability. Our approach designs a stable limit cycle ($x^*(t)$) in a robot via fast and reliable optimization. \textbf{C.} Multiple options for constraining and maximizing stability in numerical trajectory optimizations, which this paper presents and benchmarks for accuracy of stability prediction and computational speed.}
    \label{fig:limit-intro}
\end{figure}



In this paper, we cast the open-loop stability problem as a single-stage trajectory optimization problem with explicit stability constraints, and we transcribe it into a nonlinear program using Direct Collocation, which is a fast, reliable, and scalable technique for nonlinear optimization \cite{betts1998survey}\cite{kelly2017introduction}. By allowing for stability to be explicitly formulated, our approach eliminates the need to think up a clever way to encode stability implicitly for a given system -- e.g., in terms of the post-impact states for bipedal robots -- making it applicable to a broad range of dynamical systems. Furthermore, because limit cycle stability is defined using eigenvalues, which we demonstrate can be bounded rigorously -- both loosely and tightly -- our approach provides flexibility in balancing trade-offs between the speed and accuracy of the optimization solution. In fact, the eigenvalue formulations we use extend beyond limit cycle stability to other asymptotic stability criteria. For example, one can even obtain an asymptotically unstable fixed point of $\dot{x} = f(x)$ by constraining the eigenvalues of the linearized flow $\partial f/ \partial x$ to the open right-half plane using the formulations we propose. This applicability to a variety of dynamical systems and notions of stability, the ease with which these notions can be defined as constraints/objectives, and the computational efficiency of our single-loop optimization strategy paired with the analytical tractability of direct collocation, sets our approach apart from existing methods.


\section{Theoretical Background}\label{poincare}
A \textit{limit cycle} is a closed trajectory in state space that has no neighbouring closed trajectories \cite{steven1994nonlinear}. Finding these cycles is important to the design of legged robots, as stable limit cycles find extensive use in engineering stable robotic gaits \cite{dai20132}\cite{hobbelen2007disturbance}\cite{hubicki2015limit}. In Section \ref{problem-formulation}, we discuss our methods for identifying stable limit cycles. For now, let us formally define limit cycle stability by following Poincaré’s approach. 

\noindent Consider a differential equation
\begin{align} \label{eq1}
    \frac{dx}{dt}=f(x),
\end{align} 
where $x: [0, \infty) \to \mathbb R^n \text{, s.t. } x \in \bigcup_{k \geq 2} C^k$, is a periodic solution that can be thought of as originating from a codimension-one surface $\mathcal P$ -- a \textit{Poincar\'e section} -- of the $n$-dimensional phase space $\mathbb{R}^n$. The surface $\mathcal P$ is set up so that the flow cuts it transversely, i.e., the normal to $\mathcal P$ at $x_0\coloneqq x(0)$, $n_{\mathcal P}(x_0)$, satisfies $\langle n_{\mathcal P}(x_0) \,, f(x_0) \rangle \neq 0$, where $\langle\cdot,\cdot\rangle$ denotes the inner product on $\mathbb R^n$. Since $x$ is a periodic solution to (\ref{eq1}), there exists $\tau: \mathbb{N} \to \mathbb{R}^+$ such that if $x_k = x(\tau(k))$, then $x_k \in \mathcal P$. The $(n-1)$-dimensional discrete dynamical system $\phi_{\mathcal P}: \mathcal P \to \mathcal P$ such that $\phi_{\mathcal P}(x_{k-1}) = x_k$, called the \textit{Poincar\'e map}, encodes the transverse dynamics that describe how the trajectories ``near'' $x$ behave.


Say $x^*: [0, \infty) \to \mathbb R^n$ is a periodic solution to (\ref{eq1}) such that $x^*_0 \in \mathcal P$ and $x_{k}^* = \phi(x_{k-1}^*) = x_{k-1}^*$. Then, $x^*_k$ is a \textit{fixed point} of the Poincar\'e map $\phi_{\mathcal P}$, and we say that it is stable inside a neighbourhood $\mathcal U \subset \mathcal P$ if for all periodic solutions $x$ to (\ref{eq1}) such that $x_0 \in \mathcal U$, $\lim_{k \to \infty} x_k = x^*_0$. Now, suppose $\eta_k$ is the distance between $x^*$ and a ``nearby'' trajectory $x$ at the $k^{\text{th}}$ Poincar\'e section ``intersection''. Then
\[
    \eta_{k} = \frac{\partial \phi_{\mathcal P}}{\partial x} \bigg\rvert_{x^*_0} \eta_{k-1}
\]
to the first order \cite{teschl2012ordinary}.
We can see that $x^*_0$ is stable if $\eta_k \to 0$ as $k \to \infty$, i.e., if the eigenvalues of $\partial \phi_{\mathcal P}(x_0^*)/\partial x$ lie inside the unit circle in the complex plane $\mathbb C$. Notice that if $x^*_0$ is a fixed point of the Poincar\'e map $\phi_{\mathcal P}$, then the corresponding trajectory $x^*$ is a limit cycle. One may thus define a limit cycle $x^*$ as stable if the corresponding fixed point $x^*_0$ of the discrete map $\phi_{\mathcal P}$ is stable. This definition of stability makes intuitive sense: it describes whether trajectories in a neighbourhood $\mathcal U' \subset \mathbb R^n$ of the limit cycle asymptotically converge to it. And this intuition leads to an alternate procedure for evaluating limit cycle stability. Say $x^*$ is a limit cycle of (\ref{eq1}) and $x^* + \delta x$ is a trajectory in a neighbourhood $\mathcal U'$ of $x^*$. One can see that after one period, the difference, $\delta x$, between $x^*$ and $x^* + \delta x$ is given to the first order by
\[
    \delta x(T) = \frac{\partial x^*}{\partial x_0} \bigg\vert_{T} \delta x(0),
\]
where $T$ is the period \cite{seydel2009practical}. Clearly, the matrix $\mathcal M(T) = \partial x^*(T)/\partial x_0$ governs whether the perturbation $\delta x(0)$ decays and the perturbed trajectory converges to the limit cycle. This matrix is called the \textit{monodromy matrix}, and notice that it is the solution at time $T$ to the variational equation
\begin{align} \label{eq2}
   \frac{d \mathcal M}{dt} = \frac{\partial f}{\partial x} \bigg\rvert_{x^*} \mathcal M, \quad \mathcal M(0) = I_{n \times n}.
\end{align}

Thus, to determine the stability of the limit cycle \( x^* \), one can integrate (\ref{eq2}) over the cycle period \( T \) and examine if the eigenvalues of \(\mathcal{M}(T)\) lie within the unit circle in \(\mathbb{C}\).

\section{Optimization Formulation}\label{problem-formulation}
We pose the task of finding an open-loop stable limit cycle as the following optimization problem:
\begin{align}
    & \min_{u,T, \mathcal I_k} \, \int_0^T \! J(x(t), u(t)) \, dt\\
    & \text{s.t. } \forall \, t \in \mathcal I_k \quad \dot{x}(t) = f_k(x(t), u(t), \psi_k),\\
    & \quad \quad \quad \quad \,\, \quad\,\,\, g(x(t), u(t)) \equiv 0,\\
    & \quad \quad \quad \quad \,\, \quad\,\,\, h(x(t), u(t)) \succcurlyeq 0,
\end{align}

\noindent where $\mathcal I_k$ implicitly defines the phases based on the phase boundaries determined by the complex-valued vector functions $g$ and $h$, where we define $g \equiv a+bi$ as $\Re(g) \equiv a$ and $\Im (g) \equiv b$, and $h \succcurlyeq a+bi$ as the relational identities $\Re(h) \geq a$ and $\Im(h) \geq b$. The functions $g$ and $h$ also include other constraints, particularly, periodicity and limit cycle stability:
\begin{align}
    & g_{lc} = x(T) - x(0)\\
    & h_{st} = {-\vert \lambda_{i}(\mathrm X) \vert} + \rho_{\text{max}} \quad \forall \, i \in \{1, \dots, n \} \label{eq:pmax_constraint}.
\end{align}
Here, \(\rho_{\text{max}}\) represents the desired upper bound on the magnitude of the eigenvalues of \(\mathrm{X} \in \mathbb{R}^{n \times n}\), which could either be the matrix \(\partial \phi(x_0^*)/\partial x\) or \(\mathcal{M}(T)\). In the case where \(\mathrm{X} = \mathcal{M}(T)\), we must also include (\ref{eq2}) as a PDE constraint for each \(f_k\). Furthermore, as discussed in Section \ref{poincare}, to ensure solutions that represent strictly stable limit cycles, one must set \(\rho_{\text{max}} < 1\) in (\ref{eq:pmax_constraint}), meaning all eigenvalues of \(\mathrm{X}\) must lie within the unit circle in \(\mathbb{C}\).

\section{Methods for Bounding Spectral Radius}\label{bounds}
\begin{figure}[t]
    \centering
    \includegraphics[scale=1.0]{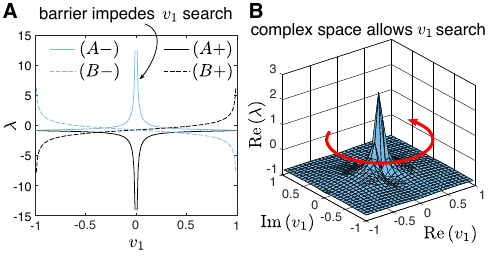}
    \caption{\textbf{A.} The constraint landspace of the $2$D eigenvalue problem\\[0.3em]
    $\phantom{.} \hspace{15pt}
        A v = \lambda v, \hspace{2pt} \text{ where } A = \begin{bmatrix} a_{11} & \hspace{-5pt} a_{12}\\[-0.3em] a_{21} & \hspace{-5pt} a_{22} \end{bmatrix} \in \mathbb R^{n \times n}, \, v = \begin{bmatrix} v_1 \\[-0.3em] v_2 \end{bmatrix} \in \mathbb R^2
    $\\[0.3em]
    The optimization variables $\lambda$ and $v_1$ here have been constrained to the real-line. It can be seen that the discontinuity at $v_1=0$ can throw off gradient-based search methods. \textbf{B.} The curve labeled $A-$ in \textbf{A} lifted to $\mathbb C^2$ from $\mathbb R^2$ by allowing $\lambda, v_1$ to be complex values, even though the $\lambda$ and $v_1$ that solve this eigenvalue problem are real-valued. The solver may be guided towards real-valued solutions through a regularization objective, 
    $
    \min_{\lambda_i} \, {-k} \left( \Im \left( \lambda_1 \right) + \Im \left(\lambda_2 \right) \right)
    $. 
    The introduction of extra dimensions to the solution space serves to condition the problem: it allows the solver to circumvent the discontinuities along the real axes by moving along the imaginary axes. However, this way of ``patching'' the CSP by lifting the constraints to a high-dimensional space does not scale well to cases where the problem is already high-dimensional.}
    \label{fig:constraints_2d}
\end{figure}

Finding solutions to the above problem is non-trivial due to the \(h_{st} \geq 0\) constraint. This section discusses the complications and presents our proposed approaches. To proceed, we define \(\mathcal{E}(X) = \{ \lambda_i(X) \}_{i=1}^{m}\) as the \textit{spectrum} of the matrix \(\mathrm{X}\), and the \textit{spectral radius}, \(\rho(\mathrm{X}) = \max \{ \vert \lambda_i(X) \vert \}_{i=1}^{m}\), as the magnitude of its largest eigenvalue. One can easily verify that to enforce \(h_{st} \geq 0\), it suffices to bound \(\rho(\mathrm{X}) \leq \rho_{\text{max}}\). However, this is difficult for general real matrices. Previous work has often assumed special structures, such as symmetry, because the eigenvalues of symmetric matrices are Lipschitz functions of their elements (Lidskii’s theorem), making semidefinite programming a viable strategy for bounding/optimizing the spectral radius \cite{lewis1996eigenvalue} \cite{lewis2003mathematics}. For non-symmetric matrices -- specifically, the matrices \(\partial \phi(x_0^*)/\partial x\) and \(\mathcal{M}(T)\) -- the spectral radius is a non-Lipschitz function of the matrix elements \cite{lewis1996eigenvalue}. Moreover, these matrices are unknown and depend nonlinearly on the optimization variables. As a result, conventional approaches to bounding the spectral radius often fail or they lead to computationally difficult constraint satisfaction problems (CSPs). For example, one can constrain $\rho(X) \leq \rho_{\text{max}}$, where $X$ can be either of $\partial \phi(x_0^*)/\partial x$ or $\mathcal M(T)$, by using a computer algebra system \cite{matlabsymb} to compute the symbolic formulation of $\text{det} \, X - \lambda I$, and adding scalar constraints to the optimization problem that bound the magnitude of the roots of this polynomial: 
$$\vert \lambda \vert \, \leq \, \rho_{\text{max}} \,\,\,\, \forall\,\lambda \,\, \text{ s.t. }  {-\epsilon} \leq \text{det} \, X - \lambda I \leq \epsilon.$$ 
However, we found that this approach scaled poorly with the size of $X$, because $\text{det} \, X - \lambda I$ gets very involved as X grows in size. To get a better conditioned formulation, we directly cast $X v = \lambda v$, where $v$ is the eigenvector corresponding to the eigenvalue $\lambda$ of $X$, as a CSP:

Find $\lambda_i \in \mathbb C$ and $v_i \in \mathbb C^n$ such that
\begin{align}
    & {-\epsilon} \leq Xv_i - \lambda_i v_i \leq \epsilon\\
    &\lambda_i - \lambda_{i-1} \geq \epsilon\\
    &{-1}+\epsilon \leq v_{i}^T v_{i-1} \leq 1-\epsilon
\end{align}
We refer to this approach as the \textit{eigenvalue definition method}. However, we discovered that this CSP was also difficult to solve, because of the discontinous nature of the solution landscape, as shown in Fig. \ref{fig:constraints_2d}A. Additionally, attempts to ``patch'' the CSP to assist the solver in navigating the solution space, as shown in Fig.~\ref{fig:constraints_2d}B, resulted in significant computational overhead in high-dimensional eigenvalue problems.

In summary, our observation has been that na\"ive general formulations of the eigenvalue problem result in ill-conditioned CSPs for constraining the spectral radius of $\partial \phi(x_0^*)/\partial x$ and $\mathcal M(T)$. In the rest of this section, we present alternative CSP formulations for bounding the spectral radius, drawing on insights from numerical linear algebra literature. In the next section, we provide a comprehensive analysis and comparison of these methods applied to the problem of swing-leg stabilization.

\subsection{Frobenius Norm Method} \label{frob-norm}
The Frobenius norm of $X$, denoted $\Vert X \Vert_F$, provides an upper bound on the squareroot of the sum of the squares of the magnitudes of the eigenvalues of $X$ \cite{horn2012matrix}:
\[
    \Vert X \Vert_F =\, \sqrt{\sum_{i =1}^n \sum_{j = 1}^n x_{ij}^2} \geq \sqrt{\sum_{i=1}^n \vert \lambda_i(X) \vert^2}
\]
Thus, bounding $\Vert X \Vert_F$ bounds  $\rho(X)$, and since $\Vert X \Vert_F$ is a differentiable function of simply the matrix elements, $x_{ij}$, the resulting CSP is well-conditioned. However, note that $\Vert X \Vert_F$ is a conservative upper bound on $\rho(X)$, and thus, in cases where $\vert \rho(X) - \rho_{\max} \vert \ll 1$, the loose Frobenius norm-based constraint may make it difficult to find a solution to the optimization problem. But outside of these cases, the Frobenius norm based bound on the spectral radius should be one's first consideration from among all the methods we propose.

\subsection{Power Iteration Method} \label{power-iter}
Power Iteration is a numerical algorithm for computing the dominant eigenvalue of a matrix, i.e., the eigenvalue $\lambda_1 \in \mathcal E(X)$ s.t. $\vert \lambda_1 \vert \,= \rho(X)$. The iteration is given by the sequence 
$$\frac{v}{\Vert v \Vert}, \frac{Xv}{\Vert Xv \Vert}, \frac{X^2 v}{\Vert X^2 v \Vert}, \frac{X^3 v}{\Vert X^3 v \Vert}, \dots, $$
which, under certain assumptions, can be shown to converge to the eigenvector $v_1$ corresponding to $\lambda_1$ \cite{vargas2020exploring}.
For computation, one can formulate this iteration as the recursion $v^{(k+1)} = {Xv^{(k)}}/{\Vert X v^{(k)} \Vert},$ which can then further be broken down into two separate recursions:
\[
    w^{(k+1)} = X v^{(k)}, \hspace{10pt}    v^{(k+1)} = {w^{(k+1)}}/{\Vert w^{(k+1)} \Vert}
\]
This recursive formulation of power iteration gives rise to the following CSP for finding the dominant eigenvalue:

Find $\lambda \in \mathbb R$, $v \in \mathbb R^n$ subject to
\begin{align}
    &v^{(n+1)} = X v^{(n)}\\
    &1-\epsilon \leq \Vert v^{(n+1)} \Vert \leq 1+\epsilon\\
    &{-\epsilon} \leq \lambda- (v^{(n+1)})^T X v^{(n+1)} \leq \epsilon
\end{align}
The superscript on $v$ indicates the node number in Direct Collocation. Note that the constraint in (13) is necessary for numerical stability in the case where $X$ is an unstable matrix, i.e., $\lim_{k \to \infty} X^k$ is undefined. However, unstable matrices are not the only limitation of the power iteration method. In particular, power iteration assumes that $\lim_{k \to \infty} {X^k v}/{\Vert X^k v \Vert}$ exists, i.e., $\lambda_1$ is strictly greater than the next ``largest'' eigenvalue $\lambda_2$ of $X$: $\vert \lambda_1 / \lambda_2 \vert >\! 1$ \cite{vargas2020exploring}. This means that the power iteration method fails if $\lambda_1 \in \mathbb C$, since $\lambda_1 \in \mathbb C$ implies $\lambda_2 = \overline{\lambda_1}$, and thus $\vert \lambda_1 / \lambda_2 \vert =\!\! 1$. However, for problems where $X$ is stable and is known to have a real dominant eigenvalue, the power iteration method can be a fast and reliable approach to bound the spectral radius. Additionally, if the design requirements for the robot necessitate the limit cycle to have real eigenvalues, the power iteration method is a suitable choice.

\subsection{Eigendecomposition Method}\label{eigendecomp} If $X$ is diagonalizable, then it can be decomposed as $X = V \Lambda V^{-1}$, where
\[
    \Lambda = \begin{bmatrix}
    \lambda_{1} \hspace{-5pt} & \hspace{-5pt} & \\[-1.2em]
    & \hspace{-5pt} \ddots & \hspace{-5pt} \\[-0.7em]
    & \hspace{-5pt} & \hspace{-5pt} \lambda_{n}
  \end{bmatrix}, \hspace{5pt}
  V = \begin{bmatrix}
    \vert \hspace{-5pt} & \hspace{-5pt} \vert & \hspace{-5pt} & \hspace{-5pt} \vert \hspace{-5pt}\\[-0.4em]
      v_1   & \hspace{-5pt} v_2  & \hspace{-5pt} \dots & \hspace{-5pt} v_n\\[-0.2em]
     \vert & \hspace{-5pt} \vert & \hspace{-5pt} & \hspace{-5pt} \vert
\end{bmatrix},
\]
are the matrices containing the eigenvalues and eigenvectors of $X$, respectively, i.e., $Xv_i = \lambda_i v_i$ \cite{trefethen1997numerical}. This eigendecomposition can be formulated as a CSP as follows:

Find $\Lambda$, $V$, $V^{-1} \in \mathbb C^{n \times n}$ such that
\begin{align}
    & -\epsilon \leq X - V \Lambda V^{-1} \leq \epsilon\\
    & -\epsilon \leq I - VV^{-1} \leq \epsilon
\end{align}

\noindent Notice that even if $X$ is a defective matrix, i.e., $\lambda_i = \lambda_j$ and $v_i = \alpha v_j$ for some $i$ and $j$, the ``wiggle room" specified by $\epsilon$ in the above CSP keeps it defined. The shortcoming of this method, however, is that the above CSP is slow to solve. This can be alleviated by restating $X = V \Lambda V^{-1}$ as $XV = V \Lambda$, thus getting rid of the constraint in (16):

Find $\Lambda$, $V \in \mathbb C^{n \times n}$ such that
\begin{align}
    -\epsilon \leq XV - V \Lambda \leq \epsilon
\end{align}

\noindent We note that the loose nature of this formulation enables the solver to ``cheat'' and give a trivial solution that sets:
$$
\Lambda = \begin{bmatrix}
    \lambda_{j} \hspace{-5pt} & \hspace{-5pt} & \\[-1.2em]
    & \hspace{-5pt} \ddots & \hspace{-5pt} \\[-0.7em]
    & \hspace{-5pt} & \hspace{-5pt} \lambda_{j}
  \end{bmatrix}, \hspace{5pt}
V = \begin{bmatrix}
    \vert \hspace{-5pt} & \hspace{-5pt} \vert & \hspace{-5pt} & \hspace{-5pt} \vert \hspace{-5pt}\\[-0.4em]
      v_j   & \hspace{-5pt} v_j  & \hspace{-5pt} \dots & \hspace{-5pt} v_j\\[-0.2em]
     \vert & \hspace{-5pt} \vert & \hspace{-5pt} & \hspace{-5pt} \vert
     \end{bmatrix},
     \hspace{5pt} j \in \{2, \dots, n\}
$$

In such a case, one may try different initial guesses for $\Lambda$ and $V$, but this trail-and-error process can be a rather significant amount of manual labour.


\subsection{Schur Decomposition Method} \label{schur-decomp}
Schur Decomposition, $X = VTV^\dagger$, where $T$ is an upper-triangular matrix, is a decomposition of a real matrix $X$ into its eigenvalues \cite{trefethen1997numerical}. This method lends itself to a widely-applicable yet relatively fast-solving CSP formulation of the eigenvalue problem:

Find $V$, $V^{-1}$, $T \in \mathbb C^{n \times n}$ subject to
\begin{align}
    & -\epsilon \leq V V^{-1} - I \leq \epsilon\\
    & -\epsilon \leq V^{-1} - V^\dagger \leq \epsilon\\
    & -\epsilon \leq X - V T V^{-1} \leq \epsilon\\
    & T_{ij} = 0 \,\, \text{ for all } j < i
\end{align}

The primary advantage of the Schur Decomposition-based formulation is its ability to provide tight bounds on the spectral radius across a broad range of dynamical systems due to its minimal assumptions. Additionally, it is relatively well-conditioned numerically and is the least sensitive to initial guesses.

\begin{figure}[tb]
    \centering
    \includegraphics[scale=1.0]{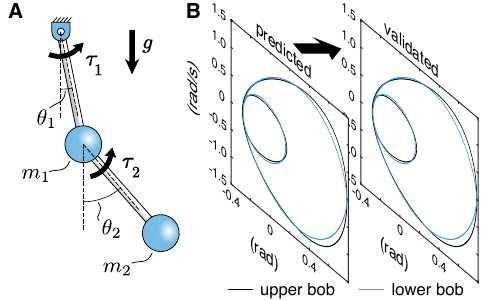}
    \caption{\textbf{A.} Robot swing-leg modeled as a double pendulum with actuated joints and viscous damping. In our experiments, we set $m_1 \!\! = \!\! m_2 \!\! = \!\! 1 \, \text{kg}$, and $L_1 \!\! = \!\! L_2 \!\! = \!\! 1 \, \text{m}$. \textbf{B.} An open-loop stable limit cycle of a moderately damped double pendulum predicted by stability-constrained optimization with an energy minimization objective. The stability of the limit cycle is validated by simulating the system from a point close to the cycle.}
    \label{fig:double_pend1}
\end{figure}

\section{Application and Analysis on Swing-leg Model}
Computing an open-loop stable limit cycle for a robot's swing-leg is a challenging control problem, making it an excellent benchmark for the stability-constrained optimization methods presented in this paper. We model the swing-leg as a double pendulum, illustrated in Fig. \ref{fig:double_pend1}A, and formulate its equations of motion as follows:
\begin{multline*}
    (m_1+m_2)L_1 \ddot{\theta}_1 + m_2 L_2 \ddot{\theta}_2 \cos(\theta_1 - \theta_2) + (m_1 + m_2) g sin\theta_1 \\+ m_2 L_2 \dot{\theta}_2^2 \sin(\theta_1 - \theta_2) + b_1 \dot{\theta}_1 = \tau_1 \quad
\end{multline*}
\vspace{-22pt}
\begin{multline*}
    m_2 L_2 \ddot{\theta}_2 + m_2 L_1 \ddot{\theta}_1 \cos(\theta_1 - \theta_2) - m_2 L_1 \dot{\theta}_1^2 \sin(\theta_1 - \theta_2) \\+ m_2 g \sin \theta_2 + b_2 \dot{\theta}_2 = \tau_2 \quad \quad \quad
\end{multline*}

These equations are derived by minimizing the action functional, $\mathcal S[\theta_1, \theta_2] := \int_{0}^{t} L ( \theta_1, \dot{\theta}_1, \theta_2, \dot{\theta}_2, t ) dt$, where $L$ is the Lagrangian of the system \cite{hand1998analytical}. 

\begin{figure}[tb]
    \centering
    \includegraphics[scale=1.0]{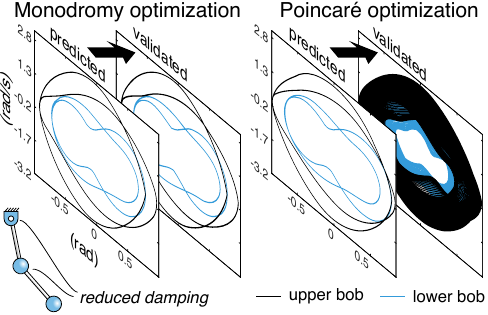}
    \caption{Stability-constrained optimization using the Poincar\'e method based definition of stability sometimes gives unstable limit cycles. This figure shows such a case, where we have a lightly damped double pendulum modeling a robot's swing-leg. \textbf{Left.} A stable limit cycle predicted by the monodromy matrix based optimization. \textbf{Right.} A stable limit cycle predicted by the Poincar\'e method based optimization. When the stability of this limit cycle is validated through simulation, we see that it is actually unstable, since the neighbouring trajectories diverge from it. }
    \label{fig:double_pend2}
\end{figure}

Through our experiments on various configurations of the swing-leg system, we have come to find the monodromy matrix method (where $X \! = \! \mathcal M(T)$) to be more reliable than the Poincar\'e method (where $X\!\!=\!\partial \phi(x_0^*)/\partial x$) for stability-constrained optimization. For the case with moderate damping ($b_1 \!\! = \!\! b_2 \!\! = \! 3.75 \, \text{Ns/m}$), both the methods performed well. In fact, we got the same limit cycle (shown in Fig. \ref{fig:double_pend1}B) from both the optimizations, and it was stable. We validated the stability of the limit cycle by simulating the swing-leg from a point close to the limit cycle and confirming that the trajectory got closer to the limit cycle over time. However, for the case with low damping ($b_1 \!\! = \!\! b_2 \!\! = \! 0.18 \, \text{Ns/m}$), shown in Fig. \ref{fig:double_pend2}, the Poincar\'e method based optimization failed: the limit cycle it gave as stable, when simulated, came out to be unstable. We believe this to be due to numerical inaccuracies in computing the Jacobian of the return map in the Poincar\'e method. Furthermore, the Poincar\'e method is sensitive to the placement of the Poincar\'e section, and a different placement of the Poincar\'e section may give a different result.

\begin{table}[b]
\centering
\begin{tabular}{@{\extracolsep{4pt}}llll}
\toprule   
{Method} & Section & {$\lambda_{\max}(\mathcal M (T))$} & {Time (s)}\\
\midrule
\textbf{Frobenius Norm} & \ref{frob-norm}  & 0.6198 + 0.2778i & 49.8\\
\textbf{Power Iteration} & \ref{power-iter} & 0.7876 + 0.0509i & 345.77\\
\textbf{Eigenvalue Definition} & \ref{bounds} & 0.8837 + 0.0958i & 287.87\\
\textbf{Eigendecomposition} & \ref{eigendecomp} & 0.7875 + 0.0493i & 242.81\\
\textbf{Schur Decomposition} & \ref{schur-decomp} & 0.7875 + 0.0493i & 73.96\\
\textbf{Symbolic} & \ref{bounds} & N/A & N/A\\
\bottomrule
\end{tabular}
\caption{Dominant eigenvalues of $\mathcal M (T)$ corresponding to the limit cycles in Fig. \ref{fig:eigval_mtds}, and the time taken to solve the optimization problem. The "N/A" in the last row shows that the optimization based on the symbolic method for eigenvalue computation was unable to converge.}
\label{tab:eig_val_comp}
\end{table}

\begin{figure}[tb]
    \centering
    \includegraphics[scale=1.0]{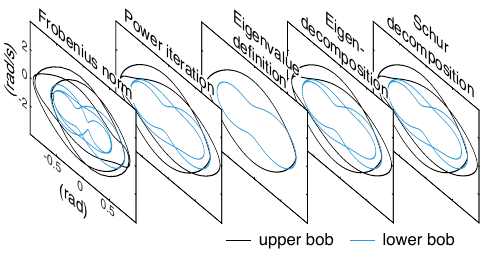}
    \caption{This figure show the limit cycles obtained from stability-constrained optimization of swing-leg using the various eigenvalue computation methods presented in this work.}
    \label{fig:eigval_mtds}
\end{figure}

In Fig. \ref{fig:eigval_mtds}, we show how the various methods we presented in Section \ref{bounds} for constraining spectral radius perform at stability-constrained optimization of the swing-leg system under low damping. All the limit cycles in Fig. \ref{fig:eigval_mtds} have been obtained against a strict stability constraint, $\rho (\mathcal M(T)) \leq 1 - \epsilon$ for a small $\epsilon$, and an energy minimization objective
\[J=\int_0^T \tau_1^2 + \tau_2^2 \, dt,\]
where $T$ is the time period of the limit cycle. The first plot in Fig. \ref{fig:eigval_mtds} shows the limit cycle obtained when the stability constraint is enforced through the Frobenius norm: $\Vert \mathcal M(T) \Vert_F \leq 1$. The first row in TABLE 1 gives the dominant eigenvalue of $\mathcal M(T)$, $\lambda_{\max}(\mathcal M(T))$, corresponding to this limit cycle. Notice that the absolute value of the dominant eigenvalue is $0.6792$, which is considerably lesser than $1 - \epsilon$, the largest spectral radius of $\mathcal M(T)$ against which the system can be stable. As we explained, this is because the Frobenius norm is a loose constraint on the dominant eigenvalue. But notice that, on the upside, the optimization problem was solved the fastest.

When the spectral radius of $\mathcal M(T)$ is bounded using the power iteration method, we get a stable limit cycle that appears very different from the one obtained using the Frobenius norm constraint, as can be seen seen in Fig. \ref{fig:eigval_mtds}. The dominant eigenvalue for $\mathcal M(T)$ corresponding to this limit cycle, as given in TABLE 1, has an absolute value of  $0.7935$, which indicates that the power iteration formulation allows for a much tighter bound on the dominant eigenvalue. Notice that the dominant eigenvalue is complex, yet the power iteration method gave a solution. This is because the imaginary part of the dominant eigenvalue is relatively small.\footnote{Even though the actual dominant eigenvalue of $\mathcal M (T)$ designed by power iteration was 0.7876 + 0.0509i, the optimization internally incorrectly calculated it to be 0.7935 + 0.0060i. This is because of power iteration's inability to work reliably with complex eigenvalues.} The power iteration method only finds solutions where $\Im \left( \lambda_{\max}(\mathcal M(T))\right) \approx 0$. This reveals a particularly useful feature of the power iteration method, as it can be used in applications where one desires real (or ``almost real'') solutions to the eigenvalue problem.

The third plot in Fig. \ref{fig:eigval_mtds} shows the limit cycle obtained by using the eigenvalue definition method. The dominant eigenvalue of $\mathcal M(T)$ for this limit cycle has a magnitude of $0.8889$. Of all the limit cycles in Fig. \ref{fig:eigval_mtds}, this one has the largest spectral radius of the associated monodromy matrix. But this behavior is not consistent across all dynamical systems and their parameterizations. 

The forth limit cycle in Fig. \ref{fig:eigval_mtds} has been obtained  using the eigendecomposition based spectral radius constraints. Notice that this is the same limit cycle as the one obtained using the Power Iteration method. The Schur decomposition method also gives this exact same limit cycle, as shown in the last plot of Fig. \ref{fig:eigval_mtds}. Despite being a tight bound, the Schur decomposition method converged the fastest after the Frobenius Norm-based method, owing to its excellent numerical convergence properties. Additionally, it was the least sensitive to initial guesses among all the methods.

These experiments indicate that different eigenvalue methods can produce various stable limit cycles. Therefore, if time permits, exploring all these methods could be valuable for designing a limit cycle that best meets design requirements.

\section{Application on Hopping Robot Model}\label{slip}
We model a hopping robot using the spring-loaded inverted pendulum (SLIP) \cite{apgar2018fast} \cite{peekema2015template}, which is described by the following hybrid dynamics:
\begin{align*}
    \ddot{z}_m &=
    \frac{k}{m} \Big( l_a-l \Big) + \frac{c}{m} \left( \dot{l}_a-\dot{l} \,\right) \! - g & \text{if } z_m = 0\\
    \ddot{z}_m &\equiv {-g} & \text{ otherwise }
\end{align*}
where $l_a$ is the input to the system describing the rate at which the actuator length changes, and $l(t) = z_m(t) - z_{\text{toe}}(t)$ is the length of the robot body.\footnote{The model's kinematic and dynamic parameters are chosen to mimic those of biological bipeds \cite{birn2014don}.} Furthermore, we specify the following task and input constraints on the model:
\begin{align*}
    &\text{(a) }\,\,\,l(t) \leq 1.1 \,\, \forall t,\\
    &\text{(b) }\,\,\,0 \leq F_{\text{net}}(t)  \leq 3 \,\, \forall t, \text{ and}\\
    &\text{(c) }\,\,\,\int_0^T F_{\text{net}}(t) \, dt  \geq  0.5, \,\, \text{ where } T \geq 1
\end{align*}

\begin{figure}[tb]
    \centering
    \includegraphics[scale=1.0]{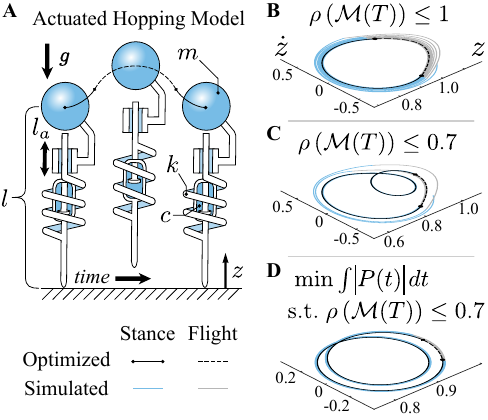}
    \caption{\textbf{A.} The hopping robot model with spring-loaded inverted pendulum (SLIP) dynamics. \textbf{B, C, D.} The result of simulating the robot from an initial state close to a limit cycle (colored black). In \textbf{B}, the limit cycle has been obtained using stability-constrained optimization with the spectral radius of the monodromy matrix constrained to less than 1. The limit cycles in \textbf{C} and \textbf{D} have been obtained by bounding the spectral radius to less than 0.7, however, in \textbf{D}, we also have an energy minimization objective.}
    \label{fig:slip_stable}
\end{figure}

Despite the challenges posed by the model's hybrid nature and frequent mode transitions, which often throw off gradient-based optimization methods, our approach effectively generates stable open-loop controllers for the hopping robot, as shown in Fig. \ref{fig:slip_stable}B-D, and that too within a relatively short time. To achieve this, we use monodromy matrix-based criteria for stability evaluation and compute the spectral radius of the monodromy matrix using Schur Decomposition.


\begingroup
\setcounter{footnote}{1}

The limit cycle shown in Fig.~\ref{fig:slip_stable}B corresponds to the case where the spectral radius of the monodromy matrix is constrained to be no larger than 1. The optimization process took less than 2 seconds to identify this limit cycle, which is stable with $\lambda_{\max}(\mathcal M(T)) ={-0.3263} + 0.7846i$.  The simulation results, depicted in the figure, confirm the stability of this limit cycle, as the system converges to it when initialized from a point in its vicinity. Fig.~\ref{fig:slip_stable}C shows the limit cycle obtained when the spectral radius of the monodromy matrix is constrained to less than $0.7$. This limit cycle has $\lambda_{\text{max}}(\mathcal M(T)) = 0.5283 + 0.3555i$. The limit cycle in Fig.~\ref{fig:slip_stable}D is also obtained against the same spectral radius constraint, however, in this case, we had an energy-minimization objective in the optimization. For this cycle, we got $\lambda_{\text{max}}(\mathcal M(T)) = 0.5180 + 0.3716i$, and the simulation indeed confirms that it is a stable limit cycle.
\endgroup

\section{Conclusion}
We framed the task of generating open-loop stable limit cycles as a single-stage trajectory optimization problem, employing two methods: the monodromy matrix method and the Poincaré method. Additionally, we tested various CSP formulations of the eigenvalue problem for bounding the spectral radius of the matrices involved. Our experiments on the robotic swing-leg model demonstrate that the monodromy matrix method is more reliable than the Poincaré method for defining limit cycle stability in optimization. We also highlight that the Schur decomposition-based formulation offers an excellent balance between numerical stability and computational efficiency, requiring minimal initial guess tuning. Additionally, our results with the hopping robot model confirm that our method performs effectively even with systems exhibiting discontinuous dynamics. Overall, our approach distinguishes itself from previous work through its straightforward applicability to a wide range of dynamical systems, computational efficiency via direct collocation, ability to accommodate various stability definitions with explicit and rigorous constraints, and capability to generate diverse stable trajectories. Further extensions of this work should incorporate dynamics with discontinuous state-space jumps into our framework to broaden its applicability.


\section*{Acknowledgement}
\noindent We extend our gratitude to Toyota Research Institute for supporting this work. We also thank the organizers and attendees of Dynamic Walking 2021, where we had the opportunity to present an extended abstract on this topic \cite{saud/stability-tractability}.

\bibliographystyle{plain}
\bibliography{refs}

\begin{thebibliography}{10}

\bibitem{apgar2018fast}
Taylor Apgar, Patrick Clary, Kevin Green, Alan Fern, and Jonathan~W Hurst.
\newblock Fast online trajectory optimization for the bipedal robot cassie.
\newblock In {\em Robotics: Science and Systems}, volume 101, page~14, 2018.

\bibitem{betts1998survey}
John~T Betts.
\newblock Survey of numerical methods for trajectory optimization.
\newblock {\em Journal of guidance, control, and dynamics}, 21(2):193--207, 1998.

\bibitem{birn2014don}
Aleksandra~V Birn-Jeffery, Christian~M Hubicki, Yvonne Blum, Daniel Renjewski, Jonathan~W Hurst, and Monica~A Daley.
\newblock Don't break a leg: running birds from quail to ostrich prioritise leg safety and economy on uneven terrain.
\newblock {\em Journal of Experimental Biology}, 217(21):3786--3796, 2014.

\bibitem{coleman1998stability}
Michael~Jon Coleman.
\newblock {\em A stability study of a three-dimensional passive-dynamic model of human gait}.
\newblock Cornell University, 1998.

\bibitem{dai2012optimizing}
Hongkai Dai and Russ Tedrake.
\newblock Optimizing robust limit cycles for legged locomotion on unknown terrain.
\newblock In {\em 2012 IEEE 51st IEEE Conference on Decision and Control (CDC)}, pages 1207--1213. IEEE, 2012.

\bibitem{dai20132}
Hongkai Dai and Russ Tedrake.
\newblock L 2-gain optimization for robust bipedal walking on unknown terrain.
\newblock In {\em 2013 IEEE international conference on robotics and automation}, pages 3116--3123. IEEE, 2013.

\bibitem{Ernst2009SpringLeggedLO}
M.~Ernst, H.~Geyer, and R.~Blickhan.
\newblock Spring-legged locomotion on uneven ground: A control approach to keep the running speed constant.
\newblock In {\em Mobile Robotics}, pages 639--644, 2009.

\bibitem{898695}
J.W. Grizzle, G.~Abba, and F.~Plestan.
\newblock Asymptotically stable walking for biped robots: analysis via systems with impulse effects.
\newblock {\em IEEE Transactions on Automatic Control}, 46(1):51--64, 2001.

\bibitem{hand1998analytical}
Louis~N Hand and Janet~D Finch.
\newblock {\em Analytical mechanics}.
\newblock Cambridge University Press, 1998.

\bibitem{hobbelen2007disturbance}
Daan~GE Hobbelen and Martijn Wisse.
\newblock A disturbance rejection measure for limit cycle walkers: The gait sensitivity norm.
\newblock {\em IEEE Transactions on robotics}, 23(6):1213--1224, 2007.

\bibitem{horn2012matrix}
Roger~A Horn and Charles~R Johnson.
\newblock {\em Matrix analysis}.
\newblock Cambridge university press, 2012.

\bibitem{hubicki2015limit}
Christian Hubicki, Mikhail Jones, Monica Daley, and Jonathan Hurst.
\newblock Do limit cycles matter in the long run? stable orbits and sliding-mass dynamics emerge in task-optimal locomotion.
\newblock In {\em 2015 IEEE International Conference on Robotics and Automation (ICRA)}, pages 5113--5120. IEEE, 2015.

\bibitem{matlabsymb}
The~MathWorks Inc.
\newblock {\em MATLAB and Symbolic Math Toolbox}.
\newblock Natick, Massachusetts, 2017.

\bibitem{kelly2017introduction}
Matthew Kelly.
\newblock An introduction to trajectory optimization: How to do your own direct collocation.
\newblock {\em SIAM Review}, 59(4):849--904, 2017.

\bibitem{lewis2003mathematics}
Adrian~S Lewis.
\newblock The mathematics of eigenvalue optimization.
\newblock {\em Mathematical Programming}, 97(1):155--176, 2003.

\bibitem{lewis1996eigenvalue}
Adrian~S Lewis and Michael~L Overton.
\newblock Eigenvalue optimization.
\newblock {\em Acta numerica}, 5:149--190, 1996.

\bibitem{mombaur2006performing}
K~Mombaur.
\newblock Performing open-loop stable flip-flops—an example for stability optimization and robustness analysis of fast periodic motions.
\newblock In {\em Fast Motions in Biomechanics and Robotics}, pages 253--275. Springer, 2006.

\bibitem{mombaur2009using}
Katja Mombaur.
\newblock Using optimization to create self-stable human-like running.
\newblock {\em Robotica}, 27(3):321--330, 2009.

\bibitem{mombaur2000stable}
Katja~D Mombaur, Hans~Georg Bock, and RW~Longman.
\newblock Stable, unstable and chaotic motions of bipedal walking robots without feedback.
\newblock In {\em 2000 2nd International Conference. Control of Oscillations and Chaos. Proceedings (Cat. No. 00TH8521)}, volume~2, pages 282--285. IEEE, 2000.

\bibitem{mombaur2005self}
Katja~D Mombaur, Hans~Georg Bock, Johannes~P Schloder, and Richard~W Longman.
\newblock Self-stabilizing somersaults.
\newblock {\em IEEE Transactions on Robotics}, 21(6):1148--1157, 2005.

\bibitem{mombaur2001human}
Katja~D Mombaur, Hans~Georg Bock, Johannes~P Schloder, and RW~Longman.
\newblock Human-like actuated walking that is asymptotically stable without feedback.
\newblock In {\em Proceedings 2001 ICRA. IEEE International Conference on Robotics and Automation (Cat. No. 01CH37164)}, volume~4, pages 4128--4133. IEEE, 2001.

\bibitem{mombaur2005open}
Katja~D Mombaur, Richard~W Longman, Hans~Georg Bock, and Johannes~P Schl{\"o}der.
\newblock Open-loop stable running.
\newblock {\em Robotica}, 23(1):21--33, 2005.

\bibitem{peekema2015template}
Andrew~T Peekema.
\newblock Template-based control of the bipedal robot atrias.
\newblock {\em Oregon State University}, 2015.

\bibitem{ringrose1997self}
Robert Ringrose.
\newblock Self-stabilizing running.
\newblock In {\em Proceedings of International Conference on Robotics and Automation}, volume~1, pages 487--493. IEEE, 1997.

\bibitem{saud/stability-tractability}
Muhammad Saud Ul~Hassan and Christian Hubicki.
\newblock Tractability of stability-constrained trajectory optimization.
\newblock In {\em Dynamic Walking 2021}, 2021.

\bibitem{seydel2009practical}
R{\"u}diger Seydel.
\newblock {\em Practical bifurcation and stability analysis}, volume~5.
\newblock Springer Science \& Business Media, 2009.

\bibitem{steven1994nonlinear}
H~Strogatz Steven and R~Strogatz.
\newblock Nonlinear dynamics and chaos: with applications to physics, biology, chemistry, and engineering, 1994.

\bibitem{teschl2012ordinary}
Gerald Teschl.
\newblock {\em Ordinary differential equations and dynamical systems}, volume 140.
\newblock American Mathematical Soc., 2012.

\bibitem{trefethen1997numerical}
Lloyd~N Trefethen and David Bau~III.
\newblock {\em Numerical linear algebra}, volume~50.
\newblock Siam, 1997.

\bibitem{vargas2020exploring}
Brian Vargas.
\newblock Exploring pagerank algorithms: Power iteration \& monte carlo methods.
\newblock 2020.

\bibitem{wensing2023optimization}
Patrick~M Wensing, Michael Posa, Yue Hu, Adrien Escande, Nicolas Mansard, and Andrea Del~Prete.
\newblock Optimization-based control for dynamic legged robots.
\newblock {\em IEEE Transactions on Robotics}, 2023.

\bibitem{westervelt2018feedback}
Eric~R Westervelt, Jessy~W Grizzle, Christine Chevallereau, Jun~Ho Choi, and Benjamin Morris.
\newblock {\em Feedback control of dynamic bipedal robot locomotion}.
\newblock CRC press, 2018.

\end{thebibliography}

\addtolength{\textheight}{-3cm}





\end{document}